\documentclass[11pt, a4paper, logo, copyright, nonumbering]{deepseek}
\usepackage[authoryear, sort&compress, round]{natbib}
\usepackage{dblfloatfix}
\usepackage{ulem}
\usepackage{caption}
\usepackage{listings}
\usepackage{verbatim}
\usepackage{dramatist}
\usepackage{xspace}
\usepackage{pifont}
\usepackage{multirow}
\usepackage{xltabular}
\usepackage{longtable}
\usepackage{hyperref}
\usepackage{tcolorbox}
\usepackage{tikz}
\usepackage{fontawesome5} 
\tcbuselibrary{breakable} 
\tcbuselibrary{skins,listings} 
\usepackage{amsfonts}
\usepackage{amsmath}
\usepackage{xcolor} 
\usepackage{amssymb}
\usepackage{lineno}
\usepackage{adjustbox}
\usepackage{graphicx}
\usepackage[bottom]{footmisc}

\usepackage{CJKutf8}
\usepackage{subcaption}
\usepackage{setspace}

\usepackage{dsfont}
\usepackage{array}
\usepackage{tabularx}
\usepackage{booktabs}
\usepackage{lipsum}
\usepackage{multicol}
\definecolor{promptgray}{gray}{0.9} 
\makeatletter
\def\@BTrule[#1]{%
  \ifx\longtable\undefined
    \let\@BTswitch\@BTnormal
  \else\ifx\hline\LT@hline
    \nobreak
    \let\@BTswitch\@BLTrule
  \else
     \let\@BTswitch\@BTnormal
  \fi\fi
  \global\@thisrulewidth=#1\relax
  \ifnum\@thisruleclass=\tw@\vskip\@aboverulesep\else
  \ifnum\@lastruleclass=\z@\vskip\@aboverulesep\else
  \ifnum\@lastruleclass=\@ne\vskip\doublerulesep\fi\fi\fi
  \@BTswitch}
\makeatother

\addto\extrasenglish{

}

\reportnumber{001}

\renewcommand{\today}{}

\title{\centering HiDream-I1: A High-Efficient Image Generative Foundation Model with Sparse Diffusion Transformer}

\author[*]{
HiDream.ai

}

\renewcommand{\phi}{\varphi}

\renewcommand{\epsilon}{\varepsilon}
\renewcommand{\imath}{\mathrm{i}}

\newlength{\restsubwidth}
\newlength{\restsubheight}
\newlength{\restsubmoreheight}
\newcommand{\rest}[2]{%
        \settowidth{\restsubwidth}{\ensuremath{#2}}
        \settoheight{\restsubheight}{\ensuremath{{}_{#2}}}
        \ensuremath{{#1\hskip 0.5pt}_{\vrule\kern2pt\parbox[b][%
        4pt][b]{\the\restsubwidth}{%
                        \ensuremath{{}_{#2}}}}}
        }

\begin{abstract}

  Recent advancements in image generative foundation models have prioritized quality improvements but often at the cost of increased computational complexity and inference latency. To address this critical trade-off, we introduce \textbf{HiDream-I1}, a new open-source image generative foundation model with 17B parameters that achieves state-of-the-art image generation quality within seconds. HiDream-I1 is constructed with a new sparse Diffusion Transformer (DiT) structure. Specifically, it starts with a dual-stream decoupled design of sparse DiT with dynamic Mixture-of-Experts (MoE) architecture, in which two separate encoders are first involved to independently process image and text tokens. Then, a single-stream sparse DiT structure with dynamic MoE architecture is adopted to trigger multi-model interaction for image generation in a cost-efficient manner. To support flexiable accessibility with varied model capabilities, we provide HiDream-I1 in three variants: \textbf{HiDream-I1-Full} (the full version with 50+ steps), \textbf{HiDream-I1-Dev} (guidance-distilled version with 28 steps), and \textbf{HiDream-I1-Fast} (the fastest version with only 14 steps).

  Furthermore, we go beyond the typical text-to-image generation and remould HiDream-I1 with additional image conditions to perform precise, instruction-based editing on given images, yielding a new instruction-based image editing model namely \textbf{HiDream-E1}. Ultimately, by integrating text-to-image generation and instruction-based image editing, HiDream-I1 evolves to form a comprehensive image agent (\textbf{HiDream-A1}) capable of fully interactive image creation and refinement. To accelerate multi-modal AIGC research, we have open-sourced all the codes and model weights of HiDream-I1-Full, HiDream-I1-Dev, HiDream-I1-Fast, HiDream-E1 through our project websites: \url{https://github.com/HiDream-ai/HiDream-I1} and \url{https://github.com/HiDream-ai/HiDream-E1}. All features can be directly experienced via \url{https://vivago.ai/studio}.

  \end{abstract}
  
\begin{document}
    \begin{figure}[htbp]
    \centering
    \includegraphics[width=0.95\linewidth]{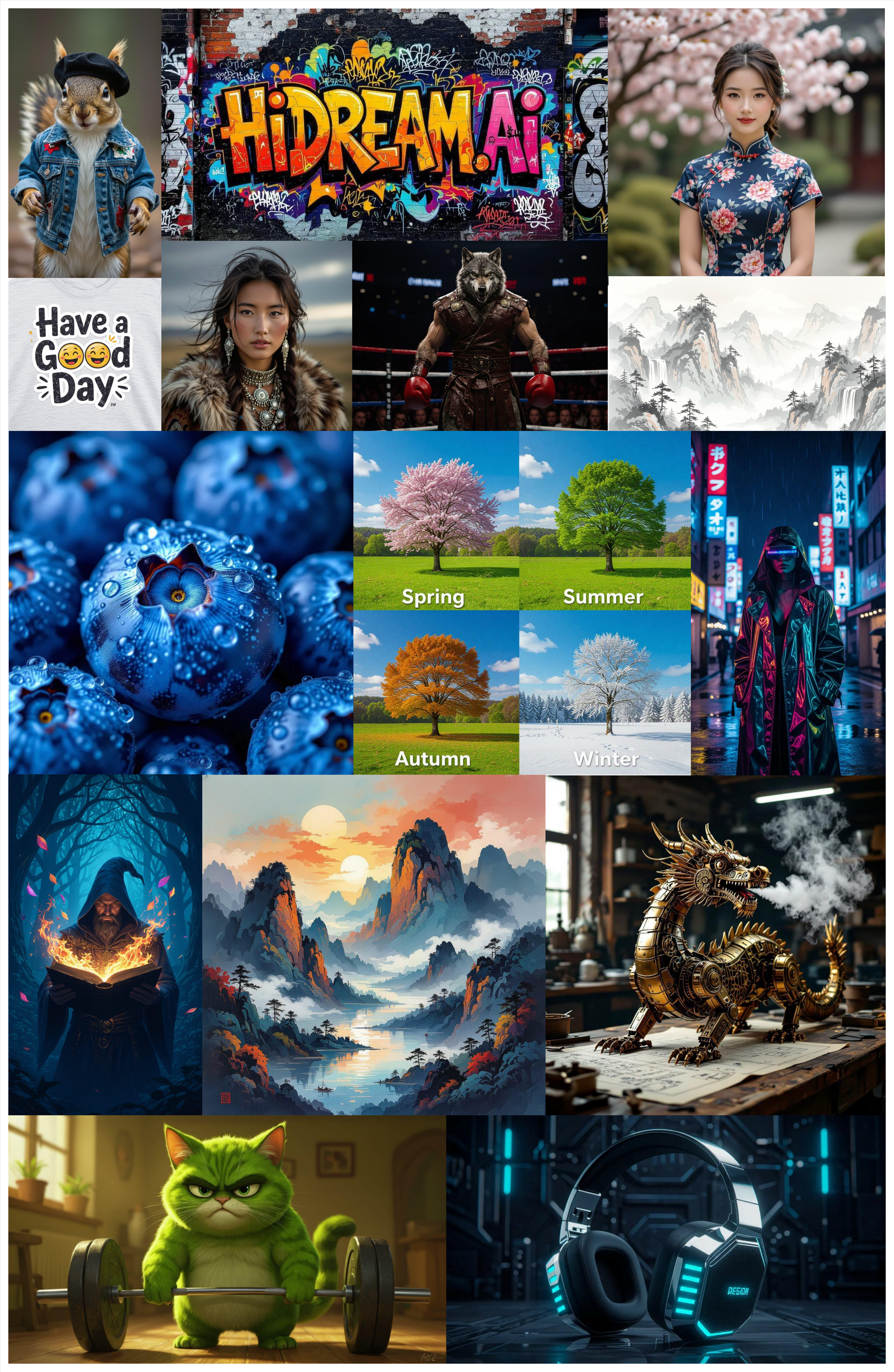} 
    \caption{Generated images by HiDream-I1.}
    \label{fig:gen_images}
  \end{figure}
  \maketitle
  
  
  \newpage
  
  \begin{spacing}{0.9}
  \tableofcontents
  \end{spacing}
  
  \newpage
  
\section{Introduction}

Recent advancements in image generative models have dramatically transformed the creative landscape, with pioneers such as Stable Diffusion, Midjourney, DALLE-3, and Imagen pushing the envelope in text-to-image generation. These models have set new benchmarks for visual fidelity and creativity, fueling applications across digital art, entertainment, design, and beyond. Despite their impressive outputs, many state‑of‑the‑art models grapple with increased computational complexity and longer inference times, making real‑time, cost‑effective deployment a persistent challenge.

To bridge this gap, we present HiDream-I1, a new open‑source image generative foundation model with 17 billion parameters that achieves state‑of‑the‑art image quality within seconds. HiDream-I1 is built on a new sparse Diffusion Transformer (DiT) structure, which first uses dual-stream DiT architecture to separately process image and text tokens and then employs single-stream DiT architecture to enable multi-modal interaction. Both dual-stream and single-stream DiT architecture are remoulded with dynamic Mixture-of-Experts (MoE) design that aims to dynamically route data through specialized expert modules based on input characteristics. In this way, the whole sparse DiT structure achieves a remarkable balance between computational efficiency and output image generation quality.

Understanding that user requirements vary widely across different applications, we offer HiDream-I1 in three distinct variants:
1) \textbf{HiDream-I1-Full}: The full‑scale model with over 50 diffusion steps, designed for scenarios where the utmost image quality is paramount.
2) \textbf{HiDream-I1-Dev}: A guidance‑distilled version optimized to operate in 28 diffusion steps, offering a well‑balanced trade‑off between quality and computational demand.
3) \textbf{HiDream-I1-Fast}: The fastest version, employing only 14 diffusion steps to deliver high-quality image generation in seconds, ideal for real-time applications.

Beyond its robust text‑to‑image synthesis capabilities, HiDream-I1 lays the foundation for interactive image manipulation (i.e., instruction-based image editing). We extend its functionality with \textbf{HiDream-E1}, an instruction-based image editing model that integrates additional image conditions to allow users to perform precise modifications through natural language commands. This advancement transforms HiDream-I1 into a versatile AIGC tool that not only generates images but also edits them iteratively, culminating in the evolution of a comprehensive image agent (namely \textbf{HiDream-A1}). This image agent facilitates fully interactive image creation and editing, setting the stage for next‑generation user experiences in generative AI.

In summary, HiDream-I1's contributions are multifaceted:
\begin{itemize}
    \item     
\textbf{Excellent Cost-Effectiveness: Sparse Diffusion Transformer Architecture.} The overall sparse DiT architecture elegantly integrates Sparse Mixture-of-Experts (MoE) technology within the DiT framework. Different expert models can learn to handle various types of text inputs, enabling the model to more accurately understand textual content and generate images with diverse styles and subjects. More importantly, Sparse DiT enhances model performance while controlling computational overhead, achieving exceptional cost-effectiveness.
    \item 
\textbf{Powerful Distillation Technology: GAN-powered Diffusion Model Distillation.}
By incorporating generative adversarial learning into diffusion model distillation, the approach leverages GANs' ability to nicely capture details and sharpen edges in image generation, which is commonly sacrificed in typical diffusion model distillation. Accordingly, this way not only distills the diffusion model but also further enhances the realism and clarity of the generated images, achieving dual optimization of speed and quality.
\item 
\textbf{Performance Advantage: Leading Benchmark Results in both human preference and semantic alignment.}
Currently, our HiDream-I1 achieves state-of-the-art performances on three benchmarks. Firstly, in the HPS benchmark \cite{}, which holistically assesses semantic relevance, image quality, and aesthetics by calculating the Human Preference Score v2.1 (HPSv2.1), HiDream-I1 attains the best performances across various styles (such as animation, concept art, painting, and photography). Additionally, in the GenEval \cite{} and DPG-Bench \cite{} benchmarks, both evaluating the semantic relevance between generated images and input text prompts, HiDream-I1 also achieves top semantic relevance, demonstrating strong prompt-following capabilities.
\item 
\textbf{Exceptional Scalability: Evolving from Image Generation to Multimodal Agents.}
Building upon HiDream-I1, we rapidly expanded to the intruction-based image editing model HiDream-E1. After that, combining text-to-image generation with interactive image editing, HiDream-I1 and HiDream-E1 achieve a "speak-and-it-shall-be" effect similar to GPT-4o's image generation and editing. Furthermore, we release a new image agent product, HiDream-A1, which integrates image generation, understanding, and interactive editing into a conversational large model. The introduction of HiDream-A1 represents an upgrade in user interaction that eliminates the need for users to switch platforms or adjust complex parameters, further lowering the barrier to using AIGC tools for creation.
\end{itemize}

\begin{figure}[ht]
  \centering
  \includegraphics[width=0.9\linewidth]{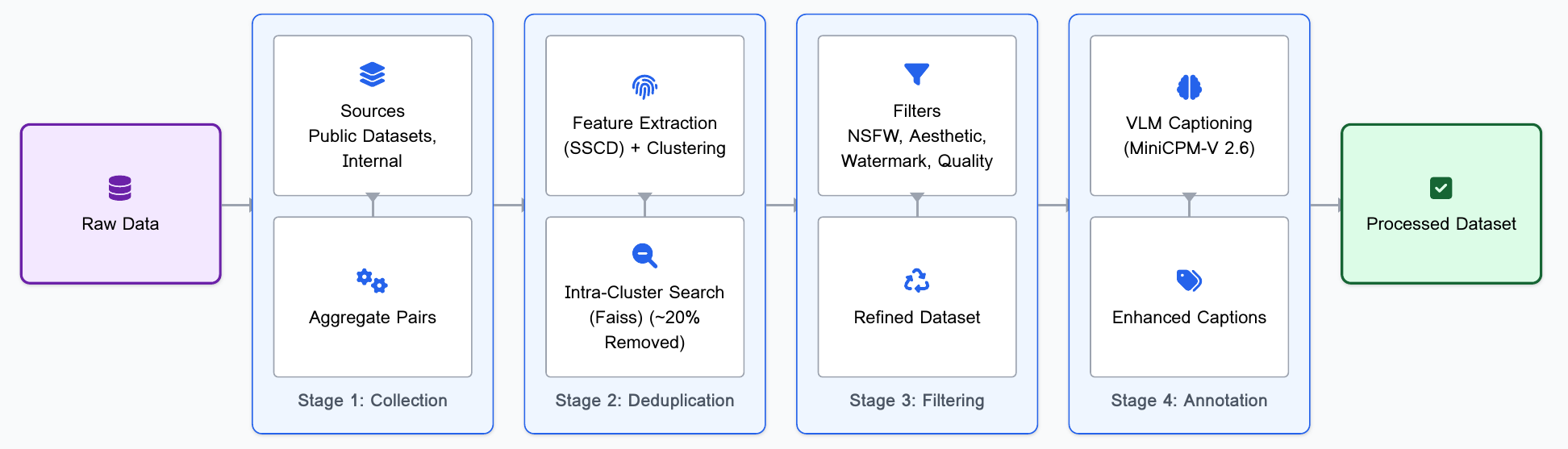} 
  \caption{Pipeline of data pre-processing.}
  \label{fig:datapipeline}
\end{figure}

\section{Data Pre-processing}

The capabilities of large-scale text-to-image models are commonly driven by the quality, diversity, and sheer scale of their training data. Thus we implemented a rigorous data curation pipeline involving systematic data collection, meticulous deduplication, comprehensive multi-faceted filtering, and detailed annotation. The overall pipeline is depicted in Figure~\ref{fig:datapipeline}.

\noindent
\subsection{Data Collection}
Our data collection strategy aimed for both breadth and diversity. We initially aggregated a massive pool of candidate data, drawing from web-sourced datasets and internal copyrighted images. For web-sourced data, we collected not only the images but also the accompanying textual information (tags, descriptions, etc.). A key focus during this initial collection stage was to intentionally capture a wide spectrum of visual content—spanning diverse artistic styles, subjects, resolutions, and aspect ratios.

\noindent
\subsection{Data Deduplication}

To enhance training efficiency and mitigate the risk of model memorization \cite{somepalli2023diffusion}, we implemented a comprehensive deduplication process designed to remove both exact duplicates and visually similar images. Given the dataset's scale, performing direct all-pairs comparisons was computationally prohibitive. Therefore, we adopted an efficient two-stage approach:
\begin{enumerate}
    \item \textit{Feature Extraction and Approximate Clustering.} First, we extracted robust visual features for each image using the state-of-the-art SSCD model \cite{pizzi2022self}. We then applied k-means clustering to a representative feature subset (N=2 million), partitioning the entire feature space into 16,000 distinct clusters and grouping potentially similar images together.
    \item \textit{Intra-Cluster Deduplication.} Second, within each cluster, we performed exact similarity searches using the GPU-accelerated Faiss library \cite{douze2024faiss}. Images whose feature vectors exceeded a predefined similarity threshold were identified as near-duplicates and subsequently removed from the dataset.
\end{enumerate}
This two-stage deduplication strategy led to the removal of approximately 20\% of the initially collected images, significantly reducing data redundancy.

\noindent
\subsection{Data Filtering}
Raw image collections often contain content that is low-quality, irrelevant, or otherwise detrimental to model training and safety. To address this, we applied a series of filtering steps to refine the dataset:
\begin{itemize}
    \item \textit{Content Safety Filter.} Images flagged as potentially inappropriate were identified and removed using a pre-trained Not Safe For Work (NSFW) classifier \cite{laion2024clip}.
    \item \textit{Aesthetic Quality Filter.} An aesthetic scoring model \cite{laion2024aesthetic} was employed to predict the visual appeal of each image. Images falling below a specified aesthetic score threshold were discarded.
    \item \textit{Watermark Filter.} A specialized detection model \cite{laion2024watermark} was used to identify and filter out images containing conspicuous watermarks.
    \item \textit{Technical Quality Filters.} Objective image quality metrics were used for further refinement. Images receiving low scores according to the Top-IQ quality assessment metric \cite{chen2024topiq} were removed. Images were temporarily encoded in JPEG format to calculate their bytes-per-pixel ratio. Those yielding unusually low ratios, often indicative of poor detail or excessive compression artifacts, were also filtered out.
\end{itemize}

\noindent
\subsection{Data Annotation}

Next, to generate descriptive captions at scale, we employed the MiniCPM-V 2.6 Vision-Language Model (VLM) \cite{openbmb2024minicpm} to automatically annotate each image. To enhance the specificity and factual accuracy of the generated captions, the VLM was provided with both the image content and any pre-existing metadata (such as user tags or original short descriptions) as input context.

Under specific instructions, the VLM aims to elicit detailed, objective descriptions. These instructions emphasize capturing key visual aspects—including objects, attributes, spatial relationships, and artistic style—while strictly minimizing hallucination and adhering to constraints on tense and length. Note that a new prompting strategy is leveraged to encourage the generation of captions with varying lengths to better reflect the diversity of typical user prompts.

\begin{figure}[ht]
    \centering
    \includegraphics[width=0.9\linewidth]{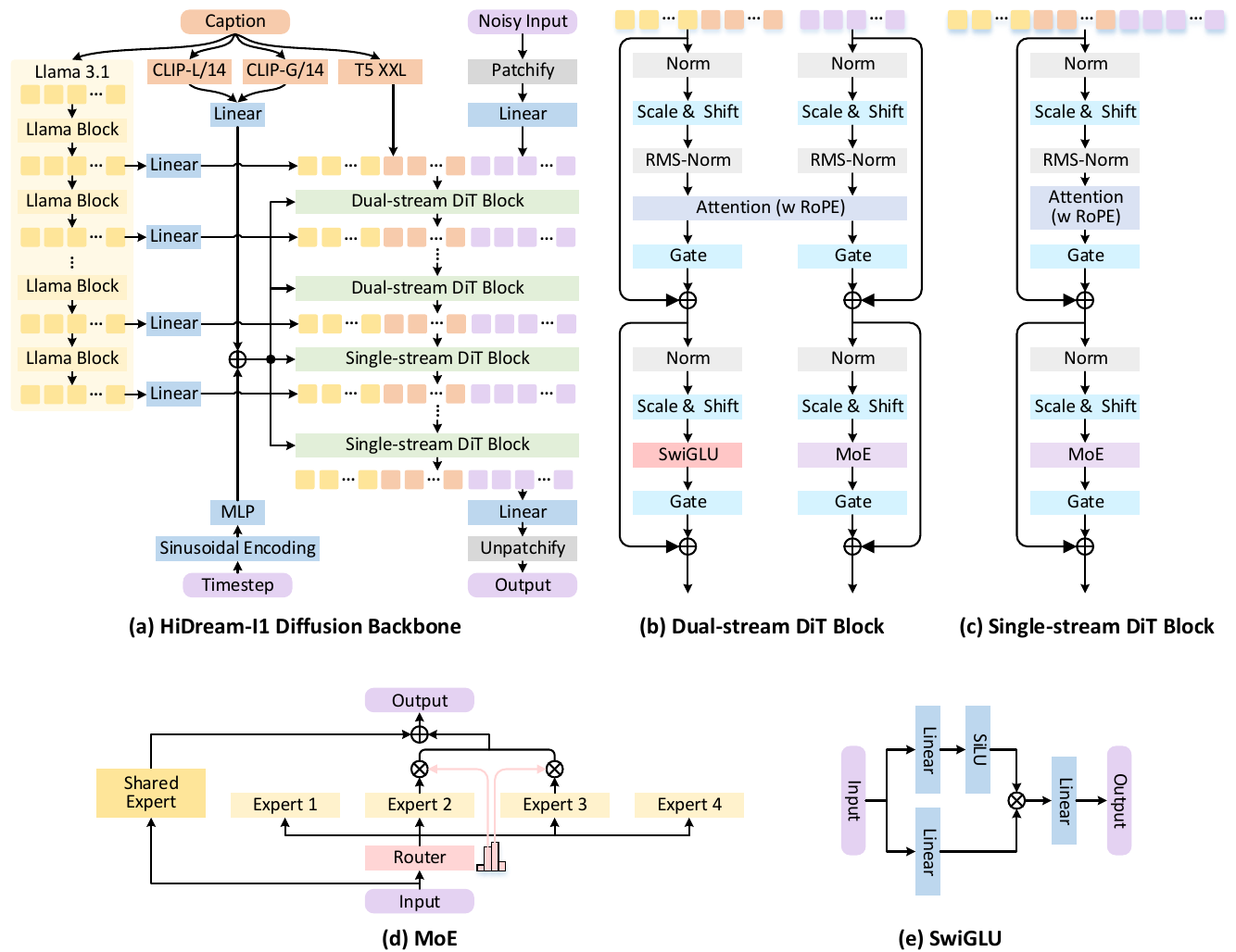} 
    \caption{Overall framework of the HiDream-I1 model.}
    \label{fig:framework}
\end{figure}

\section{Model Architecture: HiDream-I1}
\label{sec:model_architecture}

HiDream-I1 is a large-scale image generative foundation model built upon the principles of flow matching \cite{lipman2022flow} and leveraging a new sparse Diffusion Transformer (DiT) architecture. Flow matching aims to learn a continuous-time transformation from a simple prior distribution (e.g., Gaussian noise) to a complex data distribution (target images) by modeling a velocity field. Specifically, we define a path between a noise sample $X_0 \sim \mathcal{N}(0, \mathbf{I})$ and a target image $X_1$. A common choice, used here, is the linear interpolation path $X_t = (1 - t) \cdot X_0 + t \cdot X_1$ for a time $t \in [0, 1]$. The corresponding constant velocity field is $V_t = X_1 - X_0$. Our model, parameterized by $\theta$, predicts this velocity field, $u(X_t, y, t; \theta)$, conditioned on the noisy image $X_t$, the text embedding $y$, and the timestep $t$. The training objective is to minimize the mean squared error between the predicted and target velocities:
\begin{equation}
    \mathcal{L}_{\text{FM}} = \mathbb{E}_{t,X_0,X_1,y} \left[ \|u(X_t, y, t; \theta) - (X_1 - X_0)\|^2 \right]
    \label{eq:fm_loss}
\end{equation}
The core architecture, depicted in Figure~\ref{fig:framework}, comprises two main modules: a hybrid text encoding module and a unique DiT backbone incorporating sparse Mixture-of-Experts (MoE). Similar to previous works such as SD3 \cite{esser2024scaling} and FLUX \cite{blackforestlabs}, HiDream-I1 operates on latent spaces.

\subsection{Hybrid Text Encoding}
\label{subsec:text_encoder}

The ability to accurately translate textual prompts into effective conditioning signals is paramount for high-fidelity text-to-image synthesis. Recognizing that different text encoder architectures possess complementary strengths, HiDream-I1 employs a hybrid strategy integrating representations from four distinct sources:

\begin{enumerate}
    \item \textbf{Long-Context CLIP.} We utilize extended-context versions of CLIP-L/14 and CLIP-G/14 \cite{zhang2024long}, which can process longer token sequences than standard CLIP models. These provide robust, globally-aware visual grounding embeddings, presumably pooled into a single vector $h_{\text{clip}} \in \mathbb{R}^{d}$, used for global conditioning via adaptive layer normalization (adaLN).
    \item \textbf{T5 Encoder.} A T5-XXL encoder is incorporated for its proficiency in parsing complex textual structures, yielding a sequence of contextual token embeddings $h_{\text{t5}} \in \mathbb{R}^{M \times d}$.
    \item \textbf{Decoder-only LLM.} To capture deep semantic understanding, we leverage a powerful decoder-only large language model (Llama 3.1 8B Instruct). Crucially, features are extracted from multiple intermediate layers of the LLM, $h_{\text{llm}} \in \mathbb{R}^{L \times M \times d}$ (where $L$ is the number of layers tapped, $M$ is the sequence length, and $d$ is the feature dimensionality), preserving fine-grained semantic details often diluted in final-layer outputs.
\end{enumerate}

As illustrated in Figure~\ref{fig:framework} (a), the sequence embeddings from T5 ($h_{\text{t5}}$) and the selected intermediate LLM layers ($h_{\text{llm}}$) are processed (e.g., via linear projections) and concatenated to form the primary textual conditioning sequence $T \in \mathbb{R}^{N_T \times d}$ fed into the DiT backbone. This multi-source integration provides a rich, comprehensive textual representation guiding the generation process.

\subsection{Sparse Diffusion Transformer (DiT) Backbone}
\label{subsec:dit}

HiDream-I1 employs a novel DiT architecture on patchified latent representations. It begins with a dual-stream structure for independent modality processing, and subsequently transitions to a single-stream architecture. Notably, sparse Mixture-of-Experts (MoE) processing is integrated into both the initial dual-stream and the later single-stream stages.

\paragraph{Dual-stream DiT Blocks.}
The initial blocks of the DiT adopt a dual-stream structure, inspired by MMDiT \cite{esser2024scaling}, as shown in Figure~\ref{fig:framework} (b). Latent patch tokens $I_l \in \mathbb{R}^{N_I \times d}$ (derived from the patchified noisy input $X_t$) and the combined text tokens $T \in \mathbb{R}^{N_T \times d}$ (from the hybrid encoder) are processed in parallel pathways within these blocks. This allows for specialized initial feature extraction for each modality before interaction in the attention mechanism.

\paragraph{Single-stream DiT Blocks.}
Following a series of dual-stream DiT blocks ($L_{dual}$), the architecture transitions into a single-stream configuration. At this juncture, the image tokens and text tokens outputted by the final dual-stream layer ($I_{L_{dual}}, T_{L_{dual}}$) are concatenated along their sequence dimension. All subsequent transformer blocks ($l \ge L_{dual}$) then operate directly on this combined sequence. These single-stream blocks, like the preceding dual-stream blocks, incorporate a sparse Mixture-of-Experts structure for their feed-forward computations.

\paragraph{Sparse Mixture-of-Experts (MoE).}
Both the dual-stream and single-stream DiT blocks in HiDream-I1 utilize a sparse Mixture-of-Experts (MoE) structure, as detailed in Figure~\ref{fig:framework} (c, d), in place of a standard dense feed-forward network (FFN). Within each MoE block, a lightweight gating network, referred to as a router, dynamically directs each input token to a selected small subset of specialized FFN ``experts''. As depicted in Figure~\ref{fig:framework} (d), this includes routing to a designated shared expert as well. The individual expert FFNs employ the SwiGLU  \cite{shazeer2020glu} (shown in Figure~\ref{fig:framework} (e)). This strategy of sparse activation through MoE allows for a significant expansion of the model's capacity while concurrently maintaining computational efficiency relative to equivalently sized dense models.

\paragraph{Conditioning and Stability.}
Global conditioning from the pooled Long-CLIP features ($h_{\text{clip}}$) and the sinusoidal timestep embedding are injected into each transformer block. This is achieved via adaptive layer normalization (adaLN), modulating the scale and shift parameters applied to the tokens (the ``Scale \& Shift'' layers in Figure~\ref{fig:framework} (b, c)). Additionally, QK-normalization \cite{esser2024scaling} is applied within the self-attention mechanisms to enhance training stability.


\section{Model Training Strategy}
\label{sec:training}

Generally, HiDream-I1 is trained in a multi-stage strategy, which first builds foundational generative capabilities within a compressed latent space and then refines the model with enhanced quality, alignment, and user preference. This process leverages Latent Flow Matching and comprises two primary phases: multi-resolution pre-training and subsequent post-hoc alignment tuning.

\subsection{Pre-training Stage}
\label{subsec:pretraining}

The primary objective of the pre-training phase is to train the sparse DiT backbone (Section~\ref{subsec:dit}) to model the flow from noise to complex image representations within the latent space.

\begin{itemize}
    \item \textbf{Latent Space Operation.} We utilize a pre-trained VAE \cite{blackforestlabs} with an encoder $\mathcal{E}$ and decoder $\mathcal{D}$. During data preparation, all training images $X_1$ are first encoded into latent representations $Z_1 = \mathcal{E}(X_1)$. To accelerate training, these latent representations $Z_1$ are pre-computed and stored.
    \item \textbf{Progressive Resolution Training.} Training begins with latent codes $Z_1$, which are derived from images resized to the largest size fitting within a $256 \times 256$ resolution. The training runs for 600,000 steps. The batch size is 24 per GPU. The original images maintain their aspect ratio during resizing. Subsequently, the model weights initialize training using latents from images resized within $512 \times 512$ for 200,000 steps with batch size 8 per GPU, followed by a final stage using latents from images resized within $1,024 \times 1,024$ for 200,000 steps (batch size: 2 per GPU).
    \item \textbf{Optimization.} The model is trained using the AdamW optimizer \cite{loshchilov2017decoupled}. We set the learning rate to $0.0001$ with 1,000 linear warmup and decrease it when necessary. Fully Sharded Data Parallel (FSDP) \cite{paszke2019pytorch}, mixed-precision training and gradient checkpointing are employed to enable training.  
\end{itemize}

\subsection{Post-Training Stage}
\label{subsec:posttraining}


We further leverage refinement stages to improve prompt fidelity, aesthetic quality, and preference alignment. The model is fine-tuned using a set of high-quality human-annotated image-text pairs. Images in this dataset contain higher aesthetic quality and rich structure. The text-image alignment is also verified by the human annotators. The model is fine-tuned for 20,000 steps with a learning rate of 0.00001 and a global batch size of 64.
   
\section{Inference Acceleration via GAN-powered Diffusion Model Distillation}
\label{sec:acceleration}

The iterative generation process of the fully trained HiDream-I1 model (denoted HiDream-I1-Full) typically requires approximately 50 sampling steps, potentially limiting real-time applications. To alleviate this issue, we employ knowledge distillation techniques \cite{yin2024improved} to create significantly faster variants with reduced sampling steps. Specifically, we distill the full model into two accelerated versions: HiDream-I1-Dev, targeting 28 steps, and HiDream-I1-Fast, targeting 16 steps.

The diffusion model distillation process trains the student models (HiDream-I1-Dev and HiDream-I1-Fast) to approximate the generative function of the teacher model (HiDream-I1-Full) with fewer sampling steps. We employ DMD \cite{yin2024improved} as the primary distillation objective ($\mathcal{L}_{\text{DMD}}$). DMD aims to align the student model's predicted trajectory distribution with that of the teacher, enabling stable generation with fewer step sizes.

To maintain high perceptual quality and sharpness in the accelerated models, we supplement the DMD loss with an adversarial training objective ($\mathcal{L}_{\text{adv}}$). The student model (generator) is trained concurrently against a discriminator network. This discriminator assesses the realism of images decoded ($\mathcal{D}$) from the student's generated latents compared to real images. It utilizes multi-level features extracted by the frozen teacher backbone to trigger its classification. The final training objective of this GAN-powered distillation is a weighted sum of the two losses: $\mathcal{L}_{\text{total}} = \mathcal{L}_{\text{DMD}} + \lambda_{\text{adv}} \mathcal{L}_{\text{adv}}$.


\section{Extension for Image Editing: HiDream-E1}
\label{sec:editing_extension}
Next, we illustrate how to extend our HiDream-I1 framework to support instruction-based image editing, enabling modifications to a source image ($X_S$) based on a textual editing instruction ($y$).

To facilitate editing that is visually grounded in the original image, we leverage an approach that provides strong ``in-context'' visual conditioning. During fine-tuning, both the source image $X_S$ and the ground-truth target image $X_T$ (the result derived from applying instruction $y$ to $X_S$) are encoded into the VAE latent space ($Z_S, Z_T$). These latent maps are spatially concatenated side-by-side. The model is then trained using latent flow matching to generate the target latent $Z_T$, conditioned jointly on the source latent $Z_S$ (the visual context) and the editing instruction $y$.

To encourage the model to focus learning on relevant modifications dictated by the instruction, we employ a spatially weighted loss function ($\mathcal{L}_{\text{Edit-Weighted}}$). This loss formulation assigns greater importance (higher weight) to errors calculated in latent spatial regions where the target $Z_T$ significantly differs from the source $Z_S$. This prioritizes accurate generation of the edited areas while helping to preserve the unchanged visual context provided by the source image.

The instruction-based image editing model, termed HiDream-E1, was trained by fine-tuning the pre-trained HiDream-I1 model on a substantial dataset of 5 million (source image, editing instruction, target image) triplets. At inference time, the user provides the source image and the editing instruction, allowing the model to generate the appropriately modified image by leveraging the learned in-context editing skill.

\begin{figure}[htbp]
  \centering
  \begin{subfigure}{1\textwidth}
    \centering
    \includegraphics[width=\linewidth]{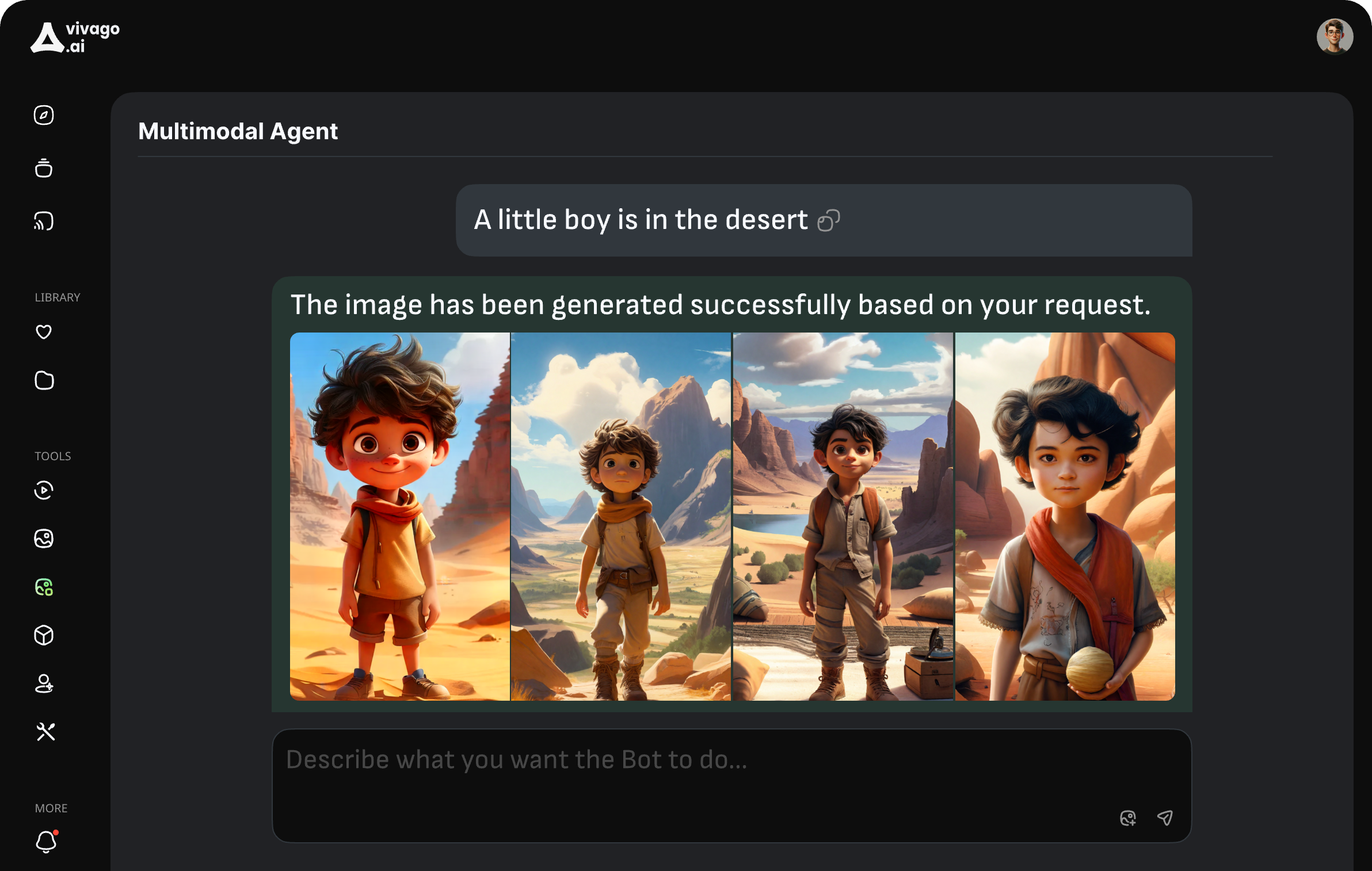}
    \caption{Text-to-image generation via image agent.}
    \label{fig:subfig_a}
  \end{subfigure}
  \hfill 
  \begin{subfigure}{1\textwidth}
    \centering
    \includegraphics[width=\linewidth]{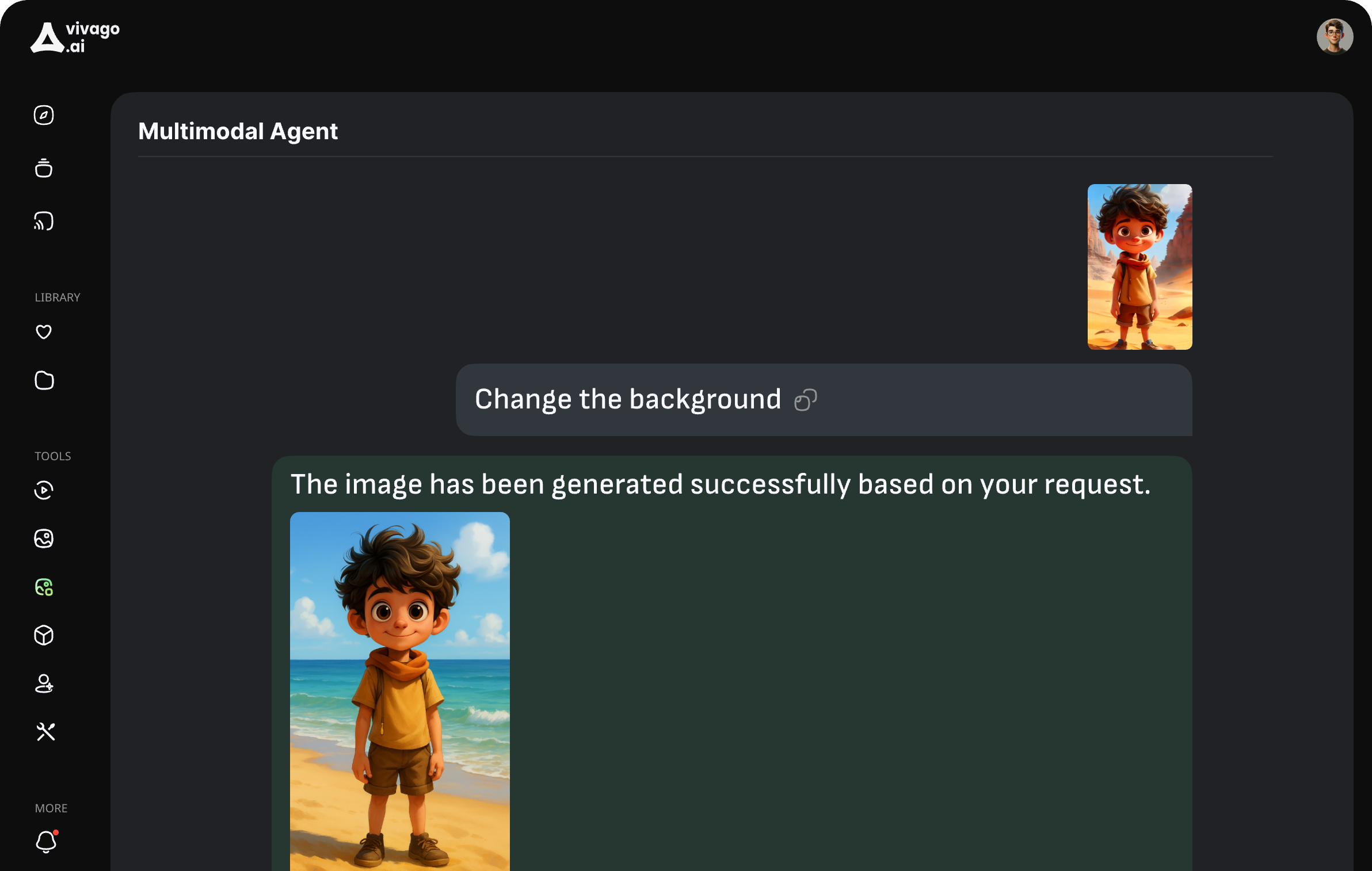}
    \caption{Instruction-based image editing via image agent.}
    \label{fig:subfig_b}
  \end{subfigure}
  \caption{Interface of image agent on vivago.ai.}
  \label{fig:agent}
\end{figure}

\section{Extension for Image Agent: HiDream-A1}
\label{sec:agent_extension}

  

Here we further introduce our new Image Agent (namely HiDream-A1) on vivago.ai, which is a unified multimodal agent system designed to seamlessly integrate text-to-image generation, image editing, and interactive understanding within a conversational AI interface. It enables users to perform complex visual content creation and manipulation using natural language dialogue. Figure \ref{fig:agent} illustrates the interface of our image agent, and two examples of text-to-image generation and instruction-based image editing tasks executed by this agent. Specifically, the process of image agent begins with User Input, which supports both natural language and visual inputs. This input is then received by a Coordinator module, the central hub that manages the whole workflow. The Coordinator determines whether the task involves Generation (generating/editing images) or Chat (a conversational interaction). If the task is Generation, the Planner module comes into play to strategize the necessary steps to fulfill the user's request. For example, depending on the specific task, the Planner might utilize an image generator (HiDream-I1) or image editor (HiDream-E1) to conduct text-to-image generation and instruction-based image editing, respectively. If the task is Chat, the agent engages in a conversational exchange with the user.

\begin{table}[t]
  \centering
  \setlength{\tabcolsep}{13pt} 
  \caption{Evaluation results on DPG-Bench. Scores represent prompt alignment accuracy (\%). Higher is better.}
  \label{tab:dpg_bench}
  \begin{tabular}{lcccccc}
    \toprule
    Model           & Overall & Global & Entity & Attribute & Relation & Other \\
    \midrule
    PixArt-alpha    & 71.11   & 74.97  & 79.32  & 78.60     & 82.57    & 76.96 \\
    SDXL            & 74.65   & 83.27  & 82.43  & 80.91     & 86.76    & 80.41 \\
    DALL-E 3        & 83.50   & 90.97  & 89.61  & 88.39     & 90.58    & 89.83 \\
    FLUX.1-dev      & 83.79   & 85.80  & 86.79  & 89.98     & 90.04    & 89.90 \\
    SD3-Medium      & 84.08   & 87.90  & 91.01  & 88.83     & 80.70    & 88.68 \\
    Janus-Pro-7B    & 84.19   & 86.90  & 88.90  & 89.40     & 89.32    & 89.48 \\
    CogView4-6B     & 85.13   & 83.85  & 90.35  & 91.17     & 91.14    & 87.29 \\
    \textbf{HiDream-I1} & \textbf{85.89} & 76.44  & 90.22  & 89.48     & \textbf{93.74} & \textbf{91.83} \\
    \bottomrule
  \end{tabular}
\end{table}

\section{Evaluation}
\label{sec:evaluation}


To comprehensively assess the generative capabilities of our HiDream-I1 and HiDream-E1, we conducted extensive evaluations against state-of-the-art text-to-image generation models and instruction-based image editing models. We employed several typical benchmarks to measure the text-to-image generation performance of HiDream-I1 in two aspects: 1) prompt adherence that reflects the model's ability to accurately understand and execute complex instructions, and 2) human preference, reflecting the visual appeal of the generated images based on predicted human preference. For the instruction-based image editing model, HiDream-E1, we evaluate both the accuracy of the edits made according to instructions and the preservation of image areas not targeted by the edits.

\subsection{Prompt Adherence Evaluation of HiDream-I1}
\label{subsec:eval_composition}

Here we evaluate the text-to-image model's capacity to accurately interpret complex prompts and synthesize corresponding images with correct object counts, attributes, relationships, and positions. Two benchmarks are adopted: DPG-Bench \cite{hu2024ella}, which provides a fine-grained assessment of prompt following across various linguistic aspects, and GenEval \cite{ghosh2023geneval}, which specifically targets challenging compositional generation tasks.

Table~\ref{tab:dpg_bench} shows the performance comparison on DPG-Bench. In general, HiDream-I1 achieves the highest overall score (85.89), indicating superior general prompt adherence capacity against existing state-of-the-art models. Specifically, HiDream-I1 shows particular strength in interpreting complex relationships (Relation: 93.74) and other detailed instructions (Other: 91.83), alongside competitive entity and attribute generation.

On the GenEval benchmark (Table~\ref{tab:geneval}), HiDream-I1 again ranks first on the overall metric (0.83). It excels across most compositional tasks, achieving the best single-object accuracy (1.00), two-object accuracy (0.98), and state-of-the-art results in counting (0.79), color generation (0.91), and color attribution (0.72). The strong performances on GenEval highlight the model's advanced capacity for compositional synthesis, accurately rendering scenes with multiple interacting elements and precise attributes.

\begin{table}[t]
  \centering
  \caption{Evaluation results on GenEval. Scores represent accuracy (\%). Higher is better.}
  \label{tab:geneval}
  \begin{tabular}{lccccccc}
    \toprule
    Model        & Overall & Single Obj. & Two Obj. & Counting & Colors & Position & Color Attr. \\
    \midrule
    PixArt-alpha & 0.48    & 0.98        & 0.50     & 0.44     & 0.80   & 0.08     & 0.07 \\
    SDXL         & 0.55    & 0.98        & 0.74     & 0.39     & 0.85   & 0.15     & 0.23 \\
    FLUX.1-dev   & 0.66    & 0.98        & 0.79     & 0.73     & 0.77   & 0.22     & 0.45 \\
    DALL-E 3     & 0.67    & 0.96        & 0.87     & 0.47     & 0.83   & 0.43     & 0.45 \\
    CogView4-6B  & 0.73    & 0.99        & 0.86     & 0.66     & 0.79   & 0.48     & 0.58 \\
    SD3-Medium   & 0.74    & 0.99        & 0.94     & 0.72     & 0.89   & 0.33     & 0.60 \\
    Janus-Pro-7B & 0.80    & 0.99        & 0.89     & 0.59     & 0.90   & 0.79     & 0.66 \\
    \textbf{HiDream-I1} & \textbf{0.83} & \textbf{1.00} & \textbf{0.98} & \textbf{0.79} & \textbf{0.91} & 0.60 & \textbf{0.72} \\
    \bottomrule
  \end{tabular}
\end{table}

\begin{table}[t]
  \centering
  \caption{Evaluation results on HPSv2.1 benchmark. Scores represent predicted human preference. Higher is better.}
    \setlength{\tabcolsep}{11pt} 
  \label{tab:hpsv2}
  \begin{tabular}{lccccc}
    \toprule
    Model                & Averaged & Animation & Concept Art & Painting & Photo \\
    \midrule
    Stable Diffusion v2.0 & 26.38    & 27.09     & 26.02       & 25.68    & 26.73 \\
    Midjourney V6         & 30.29    & 32.02     & 30.29       & 29.74    & 29.10 \\
    SDXL                 & 30.64    & 32.84     & 31.36       & 30.86    & 27.48 \\
    DALL-E 3             & 31.44    & 32.39     & 31.09       & 31.18    & 31.09 \\
    SD3           & 31.53    & 32.60     & 31.82       & 32.06    & 29.62 \\
    Midjourney V5         & 32.33    & 34.05     & 32.47       & 32.24    & 30.56 \\
    CogView4-6B          & 32.31    & 33.23     & 32.60       & 32.89    & 30.52 \\
    FLUX.1-dev           & 32.47    & 33.87     & 32.27       & 32.62    & 31.11 \\
    Stable Cascade       & 32.95    & 34.58     & 33.13       & 33.29    & 30.78 \\
    \textbf{HiDream-I1}  & \textbf{33.82} & \textbf{35.05} & \textbf{33.74} & \textbf{33.88} & \textbf{32.61} \\
    \bottomrule
  \end{tabular}
\end{table}

\subsection{Human Preference Evaluation of HiDream-I1}
\label{subsec:eval_aesthetics}

Furthermore, we evaluate visual appeal using the Human Preference Score v2.1 (HPSv2.1) benchmark \cite{wu2023human}. This benchmark employs a model trained to predict human judgments of image quality across various image styles. Higher scores indicate a stronger alignment with human preferences.

As shown in Table~\ref{tab:hpsv2}, HiDream-I1 demonstrates state-of-the-art human preference, achieving the highest average HPSv2.1 score (33.82) among all evaluated models. It consistently outperforms strong competitors across diverse styles, ranking first in all individual categories: Animation (35.05), Concept Art (33.74), Painting (33.88), and Photo (32.61). This indicates a robust capability to generate images that align well with human preferences across various visual domains.

\subsection{Evaluation of HiDream-E1}
\label{subsec:eval_e1}

In this section, we evaluated HiDream-E1's instruction-based editing capabilities on the EmuEdit \cite{sheynin2024emu} and ReasonEdit \cite{huang2024smartedit} benchmarks. EmuEdit provides a broad assessment with 3,589 samples across 10 diverse editing task types, while ReasonEdit specifically targets complex instructions with 197 challenging samples. For quantitative assessment, we adopted an automated evaluation protocol using GPT-4o. Specifically, given the source image, textual instruction, and the edited image result, GPT-4o rated two aspects on a 0-10 scale (higher is better): (1) successful execution of the editing instruction, and (2) absence of over-editing (unintended modifications). The final score per sample represents the minimum of these two ratings, thereby emphasizing both edit accuracy and preservation of unchanged regions. We compared HiDream-E1 against recent open-source competitors MagicBrush \cite{zhang2023magicbrush}, UltraEdit \cite{zhao2024ultraedit}, OmniGen \cite{xiao2024omnigen} and the closed-source model Gemini-2.0-Flash.

As shown in Table~\ref{tab:emuedit_reasonedit_bench}, HiDream-E1 achieves the highest overall average score on the EmuEdit benchmark (6.40) and also secures the leading score on the challenging ReasonEdit benchmark (7.54). The results basically demonstrate its proficiency in handling complex instructions demanding reasoning capabilities. Further analysis of the EmuEdit sub-tasks reveals specific strengths, with HiDream-E1 achieving top scores in the majority of categories: Global edits (5.32), Text manipulation (6.45), Color adjustments (7.57), Style transfer (6.49), Object removal (5.99), and Local edits (6.35). It also shows highly competitive performance in Object addition (6.98) and Background manipulation (5.01). These results again validate HiDream-E1's effectiveness for precise and reliable instruction-based image editing across a wide range of tasks.

\begin{table}[t]
  \setlength{\tabcolsep}{3pt} 
  \centering
  \caption{Evaluation results on EmuEdit and ReasonEdit benchmarks. Higher is better.}
  \label{tab:emuedit_reasonedit_bench}
  \begin{tabular}{l|ccccccccc|c}
    \toprule
    & \multicolumn{9}{c|}{EmuEdit Benchmark} &  ReasonEdit \\
    \cline{2-10} 
    Model               & Global & Add         & Text        & BG    & Color       & Style       & Remove      & Local       & Average     &    Benchmark                 \\
    \midrule
    MagicBrush          & 4.06   & 3.54        & 0.55        & 3.26  & 3.83        & 2.07        & 2.70        & 3.28        & 2.81        & 1.75                \\
    UltraEdit           & 5.31   & 5.19        & 1.50        & 4.33  & 4.50        & {5.71}      & 2.63        & 4.58        & 4.07        & 2.89                \\
    OmniGen             & 1.37   & 2.09        & 2.31        & 0.66  & 4.26        & 2.36        & 4.73        & 2.10        & 2.67        & 7.36                \\
    Gemini-2.0-Flash    & 4.87   & \textbf{7.71} & {6.30}      & \textbf{5.10} & 7.30        & 3.33        & 5.94        & {6.29}      & 5.99        & 6.95                \\
    HiDream-E1          & \textbf{5.32}   & 6.98        & \textbf{6.45} & {5.01}      & \textbf{7.57} & \textbf{6.49} & \textbf{5.99} & \textbf{6.35} & \textbf{6.40} & \textbf{7.54}       \\
    \bottomrule
  \end{tabular}
\end{table}

\section{Conclusion}
\label{sec:conclusion}
In this report, we introduced HiDream-I1, a powerful open-source image generative foundation model that balances state-of-the-art quality with exceptional efficiency. HiDream-I1 leverages a new sparse Diffusion Transformer structure, and delivers high-fidelity image synthesis within seconds. We release three variants of HiDream-I1 (i.e., HiDream-I1-Full, HiDream-I1-Dev, HiDream-I1-Fast) that supports scalable deployment across various use cases, from high-end creative workflows to real-time applications. Experiments conducted on three benchmarks (GenEval, DPG-Bench, and HPSv2.1) validate the superiority of our HiDream-I1 against state-of-the-art approaches.

Beyond its core text-to-image capabilities, HiDream-I1 evolves into a broader visual intelligence system through its extensions: HiDream-E1, which enables precise, instruction-based image editing, and HiDream-Agent, an interactive image agent that unifies generation, editing, and visual understanding within a natural language interface. Together, these components form a comprehensive, multimodal framework for interactive visual content creation.

\bibliography{main}

\newpage
\appendix

\section*{Appendix}

\section{Contributions and Acknowledgments}

\definecolor{damaiblue}{RGB}{0, 0, 100}
\definecolor{damaigreen}{RGB}{0, 100, 0}
\definecolor{damaired}{RGB}{100, 0, 0}

Contributors are listed alphabetically by the first name: 

\begin{itemize}
    \item     
\textbf{Model \& Training: Qi Cai, Jingwen Chen, Yang Chen, Yehao Li, Fuchen Long, Yingwei Pan, Ting Yao, Zhaofan Qiu, Yiheng Zhang}
\item  
\textbf{Infrastructure: Fengbin Gao, Peihan Xu, Yimeng Wang, Kai Yu}
\item  
\textbf{Data \& Product: Wenxuan Chen, Ziwei Feng, Zijian Gong, Jianzhuang Pan, Yi Peng, Rui Tian, Siyu Wang, Bo Zhao}
\item 
\textbf{Corresponding Authors: Ting Yao (tiyao@hidream.ai) and Tao Mei (tmei@hidream.ai)}
\end{itemize}

\setcounter{figure}{0}
\makeatletter 
\renewcommand{\thefigure}{A\@arabic\c@figure}
\makeatother

\setcounter{table}{0}
\makeatletter 
\renewcommand{\thetable}{A\@arabic\c@table}
\makeatother

\end{document}